\documentclass[sigconf,nonacm]{acmart}

\acmISBN{978-1-4503-XXXX-X/2018/06}

\usepackage{makecell}
\usepackage{needspace}
\usepackage{graphicx}
\usepackage{balance}
\usepackage{needspace}
\usepackage{float}
\begin{document}

\title{Trend-Aware Fashion Recommendation with Visual Segmentation and Semantic Similarity}

\author{Mohamed DJILANI}
\affiliation{%
  \institution{University of Luxembourg}
  \country{Luxembourg}}
\email{mohamed.djilani@uni.lu}

\author{Nassim {ALI OUSALAH}}
\affiliation{%
  \institution{University of Luxembourg}
  \country{Luxembourg}}
\email{nassim.aliousalah@uni.lu}

\author{Nidhal Eddine CHENNI}
\affiliation{%
  \institution{University of Luxembourg}
  \country{Luxembourg}}
\email{nidhal.chenni@uni.lu}

\renewcommand{\shortauthors}{}

\begin{abstract}
We introduce a trend-aware and visually-grounded fashion recommendation system that integrates deep visual representations, garment-aware segmentation, semantic category similarity and user behavior simulation. Our pipeline extracts focused visual embeddings by masking non-garment regions via semantic segmentation followed by feature extraction using pretrained CNN backbones (ResNet-50, DenseNet-121, VGG16). To simulate realistic shopping behavior, we generate synthetic purchase histories influenced by user-specific trendiness and item popularity. Recommendations are computed using a weighted scoring function that fuses visual similarity, semantic coherence and popularity alignment. Experiments on the DeepFashion dataset demonstrate consistent gender alignment  and improved category relevance, with ResNet-50 achieving 64.95\% category similarity and lowest popularity MAE. An ablation study confirms the complementary roles of visual and popularity cues. Our method provides a scalable framework for personalized fashion recommendations that balances individual style with emerging trends. Our implementation is available at~\textit{\textcolor{blue}{\url{https://github.com/meddjilani/FashionRecommender}}}
\end{abstract}

\keywords{Fashion Recommendation, Visual Similarity, Deep Learning, Content-Based Filtering}

\maketitle

\section{Introduction}

The proliferation of e-commerce platforms and online retailing has significantly transformed the fashion industry providing consumers with an unprecedented array of choices. However, navigating these vast selections to find suitable clothing items remains challenging, prompting the rise of personalized recommendation systems. Fashion recommendation systems harness advanced technologies particularly machine learning and deep learning, to suggest products aligning closely with individual preferences, styles, and contextual needs.

Recent advancements in Convolutional Neural Networks (CNNs) \cite{o2015introduction} have enabled significant improvements in fashion recommendation systems by efficiently capturing the rich visual attributes inherent in clothing images. Leveraging pretrained models such as ResNet50 \cite{koonce2021resnet}, DenseNet121 \cite{huang2017densely}, and VGG16 \cite{simonyan2014very}, these systems extract visual embeddings that facilitate meaningful and accurate recommendations based on visual similarity. Beyond visual features, metadata such as product categories, gender, and popularity, as well as semantic relationships between fashion items, play crucial roles in refining the recommendation accuracy.

This paper presents a comprehensive fashion recommendation framework that integrates pretrained CNNs for visual feature extraction with a multi-factor content-based recommendation engine. The proposed methodology simulates realistic retail environments and user behaviors through synthetic data generation, enhancing the evaluation robustness. Our approach addresses the limitations of traditional recommendation methods by combining visual similarity, item popularity, semantic category relationships, and user-specific attributes such as gender preferences and trendiness.

The rest of the paper is organized as follows: Section 2 discusses relevant literature in fashion recommender systems, highlighting recent advancements and remaining challenges. Section 3 outlines the proposed methodology, including detailed descriptions of visual feature extraction, metadata simulation, synthetic user purchase generation, semantic similarity measures, and recommendation scoring. Section 4 describes the system implementation and evaluation, and Section 5 concludes the paper with a summary of findings and future research directions.

\section{Related Works}

Recommender systems have become increasingly significant across various industries due to their capability to enhance user experience by providing personalized suggestions tailored to individual preferences. At their core, these systems analyze patterns of user interaction to suggest items that are likely to interest users. Initially, collaborative filtering emerged as a popular technique, where recommendations are generated based on similarities between user preferences, relying heavily on historical data of user-item interactions. However, collaborative filtering struggles with issues like sparse data—where many items have few interactions—and the cold-start problem, where recommendations for new users or new items are challenging due to insufficient data \cite{guo2012resolving}.

To overcome these limitations, content-based methods were developed. These methods utilize item-specific attributes such as descriptions, categories, and visual features, as well as user profile information, to deliver more relevant recommendations. Unlike collaborative filtering, content-based approaches can effectively manage new items since recommendations depend directly on the item’s characteristics rather than solely user interactions.

Recently, the integration of multimodal representation learning—combining multiple types of data such as text and images—has shown promising results in recommender systems. For instance, Yilma and Leiva \cite{10.1145/3544548.3581477} demonstrated that combining textual descriptions and visual imagery of items could significantly enhance recommendation quality by better capturing complex relationships that single modalities might miss. Further extending this research, Yilma et al. \cite{10.1145/3565472.3592964} examined methods of integrating these multimodal features, specifically showing that early fusion strategies—where data from multiple modalities are combined at an early stage in the model—tend to outperform later fusion strategies, thereby emphasizing the advantage of joint multimodal data representation.

Multistakeholder perspectives have also become a critical factor in modern recommender systems research. Typically, recommender systems primarily focus on end-users; however, Yilma et al. \cite{Yilma2024MOSAICMM} proposed the MOSAIC framework, which also incorporates the objectives and preferences of additional stakeholders such as content providers, curators, and platform operators. This approach ensures that recommendations are not only satisfying for users but also beneficial for other involved parties, potentially increasing the practical usability and overall acceptance of recommender systems.

In addition, the consideration of context—such as location, time, and specific environmental constraints—has gained attention due to its role in improving recommendation accuracy and relevance.  Yilma et al. \cite{10.1145/3450613.3456847} introduced the concept of Cyber-Physical-Social Systems (CPSS), highlighting the importance of factoring in real-world contexts and constraints into the recommendation process. This helps to ensure that recommendations are not only theoretically relevant but practically appropriate and actionable within specific environments.

Moreover, recommender systems have demonstrated potential beyond typical commercial applications, notably in therapeutic contexts. For instance, Yilma et al. \cite{10.1145/3613904.3642636} explored using personalized visual art recommendations as therapeutic interventions to enhance psychological well-being, illustrating how personalized recommendations can significantly impact emotional and mental health outcomes.

Specifically within the fashion domain, recommendation systems have experienced considerable innovation. Patil et al. \cite{10724655} employed ResNet-50, a deep learning architecture renowned for its ability to capture intricate visual patterns, to develop a recommendation system capable of discerning essential visual attributes of fashion items like color, texture, and style. Similarly, Sivaranjani et al. \cite{10275967} utilized Convolutional Neural Network (CNN)-based visual embeddings in conjunction with K-nearest neighbors (KNN) algorithms to enhance retrieval accuracy, surpassing traditional recommendation methods. The utilization of deep learning architectures, particularly ResNet, effectively addressed common neural network issues such as vanishing gradients, making them highly effective for image-based fashion recommendation tasks.

The integration of multimodal data has further refined fashion recommendation systems. Varma et al. \cite{00000000} introduced an AI-powered system, StyleSync, which classifies clothing types and colors using deep learning and provides personalized outfit recommendations tailored to specific occasions. Their approach emphasized user-friendly interfaces and continuous refinement through user feedback, significantly improving user satisfaction and system adaptability.

Despite these advancements, fashion recommender systems continue to face challenges, particularly in balancing highly personalized recommendations with broader market trends and real-world constraints such as inventory availability and geographic proximity. Our proposed system seeks to address these gaps by integrating sophisticated visual feature extraction techniques, realistic user simulations, semantic similarity assessments, and geographic considerations, aiming to optimize both personalization and practical applicability in real-world scenarios.

\section{Proposed Methodology}

\begin{figure*}[ht]
\centering
\includegraphics[width=0.9\linewidth]{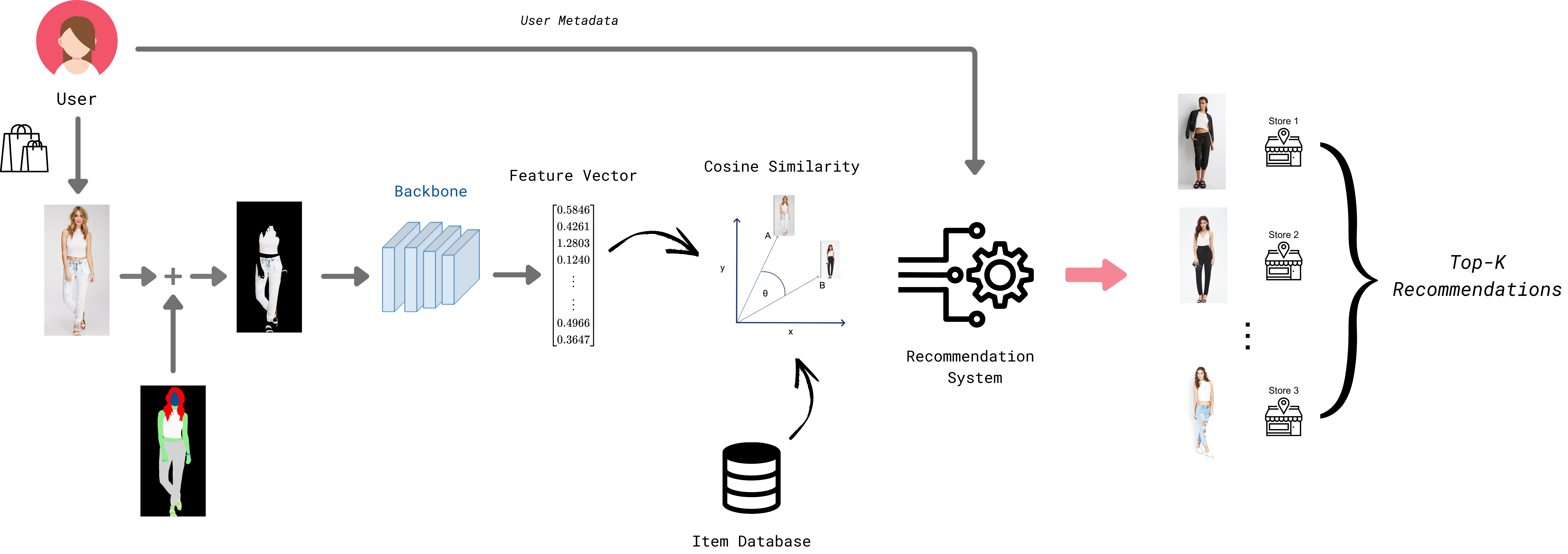} 
\caption{End-to-End Architecture of our Recommendation System, showcasing user input processing, i.e. previous bought item, feature extraction, cosine similarity computation, and generation of top-K personalized suggestions based on user meta-data.}
\label{fig:block-diagram}
\end{figure*}

We propose a content-based fashion recommendation system that leverages deep visual features, category semantics, and dynamic behavioral simulation to generate personalized product suggestions. The architecture incorporates garment-level segmentation, multiple CNN backbones, and a realistic synthetic purchase simulator to model user behavior under different fashion trends. An overview of the system is provided in Figure~\ref{fig:block-diagram}.

\subsection{Garment-Aware Visual Feature Extraction}

\begin{figure}[H]
\centering
\includegraphics[width=0.8\linewidth]{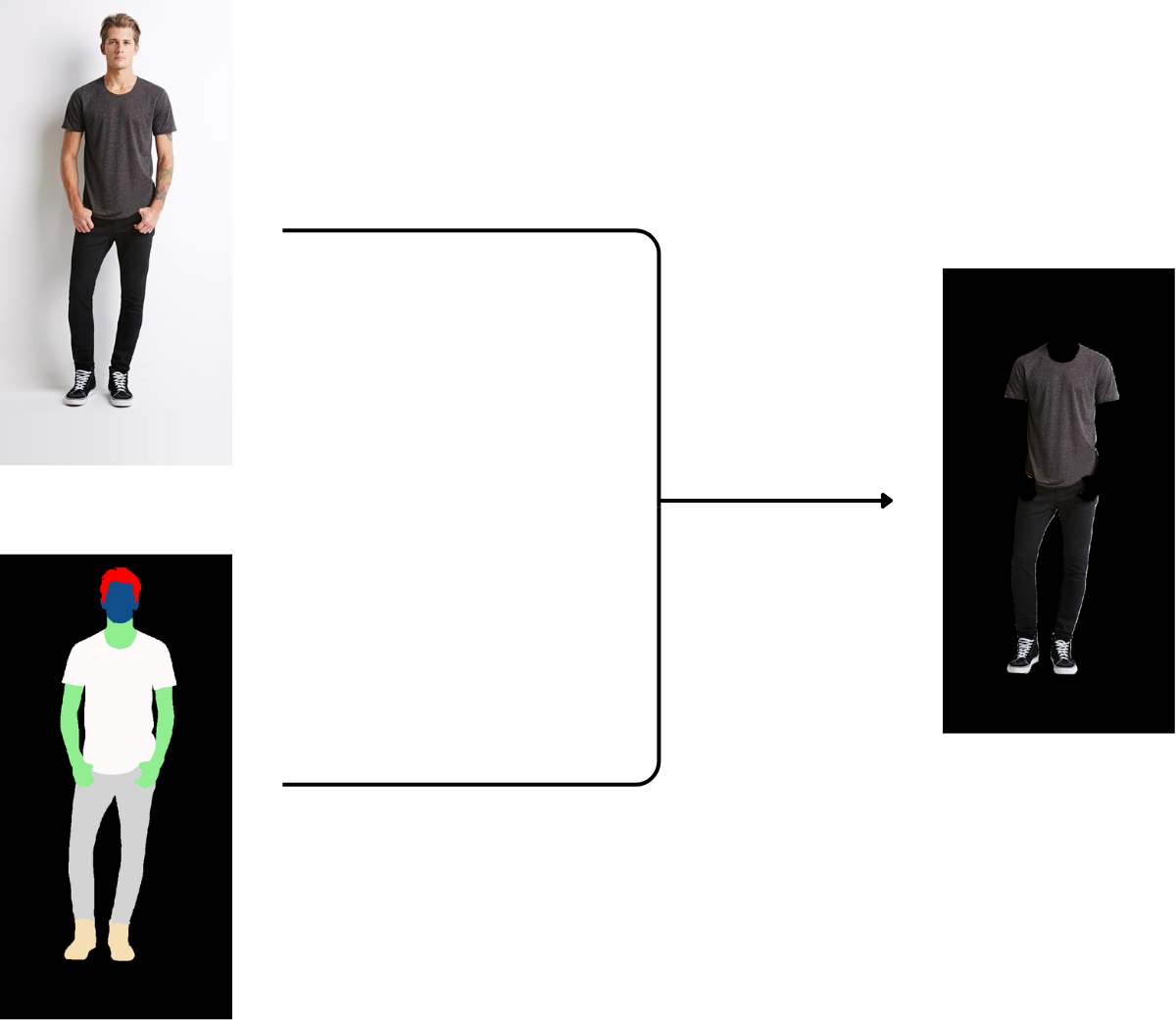} 
\caption{Overview of our Garment-Aware masking strategy for better focus on the extraction of features related to the garment only.}
\label{fig:garment-mask}
\end{figure}

To focus exclusively on clothing items, we preprocess images using semantic segmentation masks to remove non-garment regions (e.g., face, hair, skin). We apply color-based masking on segmentation outputs and extract visual features from the masked images using pretrained CNN backbones. Specifically, we employ ResNet50~\cite{he2016deep}, DenseNet121~\cite{huang2017densely}, and VGG16~\cite{simonyan2014very}. Each model outputs a 1D feature vector representing the visual appearance of a product image. An example of the our Garment-Aware masking strategy is illustated in Figure~\ref{fig:garment-mask}.

\subsection{Item Metadata and Store Assignment}

Each product image encodes metadata in its filename, including gender, category, and a unique item ID. Regular expressions are used to parse and structure this information. To simulate an omnichannel retail experience, items are assigned to synthetic store locations uniformly distributed across a geographical space. Each store is assigned random coordinates (latitude, longitude), which are later used in proximity-aware recommendation.

\subsection{Trend-Aware Synthetic Purchase Generation}

To evaluate recommendation quality under realistic behavioral patterns, we simulate the purchase history of $M$ users across one year. Each user is assigned a \textit{trendiness} score $t \in [0,1]$ reflecting their affinity for popular items. Item popularity is initialized randomly and updated dynamically as purchases occur.

The purchase history generation follows these steps:
\begin{itemize}
    \item A random item initiates the user's gender-preference.
    \item For subsequent purchases, items are sampled with a probability proportional to user trendiness and item popularity.
    \item Each transaction is timestamped with a random day in 2025.
\end{itemize}

This process ensures variability in user behavior and reflects evolving popularity patterns.

\subsection{Semantic Category Similarity}

Fashion categories are mapped to semantic groups (e.g., tops, bottoms, outerwear). We manually define inter-category similarity using a hierarchical grouping model. The similarity function assigns:
\begin{itemize}
    \item A score of 1.0 for identical categories,
    \item A score of 0.8 for different items within the same semantic group,
    \item A score in the range 0.1–0.4 for inter-group combinations based on a predefined similarity matrix.
\end{itemize}

This structured similarity enables the recommendation engine to suggest items that are not only visually aligned but also contextually relevant in terms of function and style. For instance, a user purchasing a “Cardigan” may be recommended a “Sweater” over a “Skirt,” despite potential visual similarity, due to higher semantic coherence.

\subsection{Personalized Recommendation Engine}
\label{subsec:rec_gen}

The recommendation engine generates Top-K suggestions for each user by evaluating all candidate items against their purchase history. Items already purchased are excluded, and each candidate is scored based on a combination of factors derived from both image content and behavioral simulation.

The scoring incorporates the following metrics:

\begin{itemize}
    \item \textbf{Visual Similarity:} Cosine similarity between image embeddings extracted via pretrained CNN backbones.
    \item \textbf{Popularity Score:} Normalized dynamic frequency of item purchases across all users. The Trend Penalty is an even number exponent, by increasing it, we tolerate we tolerate non big differences between the user trendiness score and the item popularity.
\end{itemize}

The relevance score is computed as:

\[
\text{Relevance} = w_v \cdot \text{VisSim} + w_p \cdot \left(1 - (\text{NormPop} - t)^{\gamma}\right) + w_c \cdot \text{CatSim}
\]

where:
\begin{itemize}
    \item $w_v$, and $w_p$are tunable weights for visual similarity, popularity modulation, and category similarity, respectively.
    \item $t$ is the user-specific trendiness score.
    \item $\gamma$ is the popularity penalty exponent.
    \item $\text{NormPop}$ is the min-max normalized popularity score of the candidate item.
\end{itemize}

Candidate items are ranked based on the final relevance score, and the Top-K highest-ranked items per purchase are returned as recommendations.

\subsection{Geographical Distance Scoring}

In addition to visual and semantic relevance, we compute the physical distance between the user's location and the store location of each recommended item. Store locations are synthetically assigned random geographic coordinates, and distances are calculated using the Haversine formula~\cite{hversine_ref} as follows,

\[
d = 2r \cdot \arcsin\left( \sqrt{\sin^2\left(\frac{\Delta\phi}{2}\right) + \cos(\phi_1) \cdot \cos(\phi_2) \cdot \sin^2\left(\frac{\Delta\lambda}{2}\right)} \right)
\]

\noindent where $r$ is the Earth’s radius (approximately 6371 km), $\phi$ and $\lambda$ are latitude and longitude values in radians, and $\Delta$ denotes the difference between user and store coordinates. While not currently used in the ranking formula, distance information is available for downstream filtering or visualization, simulating real-world user constraints such as store proximity or travel effort.

\section{Experiments}

\subsection{Experimental Setup}

To evaluate the effectiveness of the proposed recommendation system, we use the DeepFashion dataset~\cite{Liu_2016_CVPR}, a large-scale fashion benchmark containing over 800,000 clothing images annotated with category, identity, and attribute labels. We extract a subset of items annotated with gender and category, and use a consistent filename pattern to facilitate metadata parsing.

Visual features are extracted using three pretrained CNN architectures, namely, ResNet-50~\cite{he2016deep}, DenseNet-121~\cite{huang2017densely}, and VGG16~\cite{simonyan2014very}, each modified to output penultimate-layer embeddings. Images are resized to $256 \times 256$, center-cropped to $224 \times 224$, and normalized using ImageNet~\cite{russakovsky2015imagenet} statistics. The final visual features are cached for efficiency.

We simulate a one-year recommendation scenario with 50 synthetic users, each making up to 5 purchases. User preferences are influenced by a personalized \emph{trendiness score}, which modulates their likelihood to select popular items. Every item is associated with one of five simulated store locations, each randomly placed in a 2D geographical space.

\subsection{Recommendation Strategy}

For each user, we generate personalized Top-5 recommendations (with $K=5$) based on previously purchased items. Candidates are filtered by gender compatibility and scored using a weighted fusion the weighted fusion we presented earlier. The default weight configuration we use is $w_v = 0.7$, $w_p = 0.3$, though variants are tested in the ablation study. We opted for a popularity exponent of 2.

\subsection{Evaluation Metrics}

Our fashion recommendation system is ranking candidates based on a relevance score combining both visiual similarity and popularity.
To evaluate the effectiveness of our visual similarity-based fashion recommendation system, we focus on the following evaluation criteria:
\begin{itemize}
     \item  \textbf{Category Similarity:} measures whether the category of the recommended item is close to the category of the purchased item.
    \item \textbf{Gender Alignment:} measures whether the gender attribute of the recommended item matches the gender of the user who made the original purchase.
\end{itemize}

To evaluate the effectiveness of our popularity-based fashion recommendation system. We utilize:

\begin{itemize}
    \item \textbf{Popularity MAE}, we normalize then calculate the mean absolute error (MAE) between the popularity of the recommended items and the user's trendy preferences. A smaller distance indicates better effectiveness.
\end{itemize}

\subsection{Results across different Backbones}

We evaluate the impact of different CNN backbones—ResNet-50, DenseNet-121, and VGG16—on the quality of visual representations used for recommendation and how it affects the popularity MSE. Each backbone outputs a feature embedding that is used to compute cosine similarity between items. We summarize the results in Table~\ref{tab:backbone_comparison}.


\begin{table}[h]
\centering
\begin{tabular}{lccc}
\toprule
\textbf{Backbone} & \makecell{\textbf{Category}\\\textbf{Similarity}} & \makecell{\textbf{Gender}\\\textbf{Alignment}} & \makecell{\textbf{Popularity}\\\textbf{MAE}} \\
\midrule
ResNet-50~\cite{he2016deep}       & 64.95\%   & 100\%   & 12.96 \\
DenseNet-121~\cite{huang2017densely} & 63.53\%   & 100\%   & 12.47 \\
VGG16~\cite{simonyan2014very}         & 57.99\%   & 100\%   & 13.32 \\
\bottomrule
\end{tabular}
\caption{Comparison of category similarity, gender alignment, and popularity MAE across CNN backbones.}
\label{tab:backbone_comparison}
\end{table}

All models achieve perfect gender alignment (100\%), indicating reliable modeling of user gender preferences based only on visual similarity score.  

The category similarity scores for DenseNet-121 63.53\% and ResNet-50 64.95\% show an  advantage over VGG16 57.99\%, suggesting a better visual similarity. Popularity MAE values  exhibit minor variation, ranging from 21.29 for VGG16 to 21.59 for DenseNet-121.
\subsection{Ablation Study}
To isolate the contribution of Popularity and Visual Similarity in the scoring function, we perform an ablation study. We disable the scoring components by setting their corresponding weights to zero. We evaluate how the removal of these factors affects the Popularity MAE, and Category Similarity.

\begin{table}[h]
\centering
\begin{tabular}{lcc}
\toprule
\textbf{Model Variant} & \makecell{\textbf{Category}\\\textbf{Similarity}} & \makecell{\textbf{Popularity}\\\textbf{MAE}} \\
\midrule
Full Model                  &      62.16 \% &  12.92      \\
-- No Pop. Modulation          &   ↑ 63.53    \% & ↑ 31.14      \\
-- No Visual Sim.          & ↓ 40.01     \% &      ↓ 0.22\% \\
\bottomrule
\end{tabular}
\caption{Ablation study measuring the effect of removing components from the relevance scoring function.}
\label{tab:ablation_study}
\end{table}

Table~\ref{tab:ablation_study} presents the results of the ablation study evaluating the impact of different components in the relevance scoring function. As expected, removing popularity modulation (-- No Pop. Modulation) results in a substantial increase in popularity MAE, rising from 12.92 to 31.14. This configuration slightly improves the category similarity to 63.53\%, compared to 62.16\% in the full model, suggesting that the weight configuration used in the full model ($w_v=0.7$, $w_p=0.2$) is close to optimal to maximize category alignment. In contrast, disabling visual similarity (-- No Visual Sim.) leads to a drop in category similarity, down to 40.01\%, while  reducing the popularity MAE to just 0.22. This indicates that, in the absence of visual cues, the model fully prioritizes popularity signals, resulting in poor semantic matching. These findings highlight the importance of balancing visual and popularity signals to maintain personalized and relevant recommendations.

\subsection{Trendiness Preference Analysis}

We investigate the effect of user-specific trendiness preferences on the popularity and categorical similarity of recommended items. Users are stratified into three groups based on their trendiness score:

\begin{itemize}
    \item \textbf{Low Trendiness} ($t < 0.33$): Preference for niche or less popular items.
    \item \textbf{Medium Trendiness} ($0.33 \leq t < 0.66$): Balanced behavior.
    \item \textbf{High Trendiness} ($t \geq 0.66$): Preference for popular items.
\end{itemize}

We compute the average Popularity MAE and Category Similarity for each group to examine how user behavior affects recommendations. We present the results in Table \ref{tab:trendiness_effect}.

\begin{table}[h]
\centering
\begin{tabular}{lcc}
\toprule
\textbf{Trendiness Group} & \makecell{\textbf{Category}\\\textbf{ Similarity}} & \makecell{\textbf{Popularity}\\\textbf{MAE}} \\
\midrule
Low Trendiness       &   62.52 \%    &   11.84     \\
Medium Trendiness    &    65.24 \%  &    13.96    \\
High Trendiness      &     56.3 \%  &       12.05 \\
\bottomrule
\end{tabular}
\caption{Influence of user trendiness on recommendation popularity and diversity.}
\label{tab:trendiness_effect}
\end{table}

The results in Table~\ref{tab:trendiness_effect} reveal nuanced interactions between user trendiness preferences and recommendation behavior. Users with high trendiness scores—those who prefer popular items—exhibit the lowest category similarity 56.3\%, indicating that their recommendations are less aligned with their categorical interests. Conversely, users with medium trendiness show the highest category similarity 65.24\% and the highest popularity MAE 
13.96, suggesting a balanced recommendation strategy where the system still incorporates individual interests while diverging more from general popularity. Users with low trendiness, who prefer niche items, achieve relatively high category similarity 62.52 \% and the lowest popularity MAE 11.84, indicating that the model can effectively recommend niche but topically relevant items for them.

\subsection{Qualitative Case Studies}

To qualitatively assess the behavior of the recommendation engine, we present visual results for a sample user. For one of the user's purchased items, we show the Top-5 recommended items generated using each of the three backbone models. This visualization enables side-by-side comparison of visual coherence, stylistic variation, and category alignment across backbones.

\begin{figure}[h]
\centering
\includegraphics[width=0.95\linewidth]{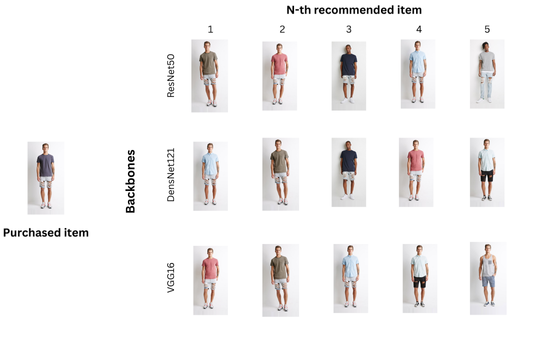} 
\caption{Top-5 recommendations for a sample user under ResNet-50, DenseNet-121, and VGG16.}
\label{fig:sample_user}
\end{figure}

These visualizations demonstrate that while all models provide reasonable suggestions, the characteristics of the recommendations differ. For example, ResNet-50 tends to produce visually similar outfits, whereas VGG16 introduces more stylistic variation. DenseNet-121 often strikes a balance between the two.

\section{Conlusion}
In this work, we simulated a social shopping environment through generating a synthetic dataset that captures user-item interactions influenced by individual trendiness preferences. This setup allowed us to explore the interplay between user behavior and recommendation dynamics in a controlled setting. We evaluated our system using both qualitative and quantitative indicators, including category similarity, gender alignment, and popularity distance, to assess the alignment between recommendations and user/item characteristics.

Our findings show that the proposed model is effective in maintaining gender alignment and achieves good performance in balancing categorical relevance and popularity, especially when trendiness preferences are moderately expressed. However, we observed limitations in visual similarity when the input consists of full-body images containing multiple garments. This suggests a potential improvement in applying object detection to isolate and compare individual items more precisely, which could improve category similarity scores and overall relevance estimation.

Looking forward, validated ground-truth signals such as gender and category alignment could be integrated directly into the relevance scoring mechanism, weighted appropriately to guide the model toward more personalized and context-aware recommendations.

\balance
\bibliographystyle{ACM-Reference-Format}
\bibliography{sample-base}
\end{document}